\documentclass[lettersize,journal]{IEEEtran}
\usepackage{booktabs}
\usepackage{multirow}
\usepackage{amsmath,amsfonts}
\usepackage{array}
\usepackage[caption=false,font=normalsize,labelfont=sf,textfont=sf]{subfig}
\usepackage{textcomp}
\usepackage{stfloats}
\usepackage{url}
\usepackage{verbatim}
\usepackage{graphicx}
\usepackage{cite}
\usepackage[linesnumbered,ruled,vlined]{algorithm2e}

\begin{document}

\title{FMDNN: A Fuzzy-guided Multi-granular Deep Neural Network for Histopathological Image Classification}

\author{Weiping Ding,~\IEEEmembership{Senior Member,~IEEE,} Tianyi Zhou, Jiashuang Huang,\\Shu Jiang, Tao Hou, and Chin-Teng Lin,~\IEEEmembership{Fellow,~IEEE}
\thanks{This work is supported in part by the National Natural Science Foundation of China under Grant 61976120, Grant 62006128, and Grant 62102199. (Corresponding author: Weiping Ding).}
\thanks{W. Ding is with the School of Information Science and Technology, Nantong University, Nantong 226019, China, and also with the Faculty of Data Science, City University of Macau, Macau 999078, China. E-mail: dwp9988@163.com.}
\thanks{T. Zhou, J. Huang, S. Jiang, and T. Hou are with the School of Computer Science and Technology, Nantong University, Nantong 226019, China. E-mails: choutyear@outlook.com; hjshdym@163.com; jshmjs45@ntu.edu.cn; houtao\_30@163.com.}
\thanks{C.-T. Lin is with the Centre for Artificial Intelligence, FEIT, University of Technology Sydney, Ultimo, NSW 2007, Australia. E-mail: chin-teng.lin@uts.edu.au.}
}

\markboth{Journal of \LaTeX\ Class Files,~Vol.~14, No.~8, August~2021}%
{Shell \MakeLowercase{\textit{et al.}}: A Sample Article Using IEEEtran.cls for IEEE Journals}


\maketitle

\begin{abstract}
Histopathological image classification constitutes a pivotal task in computer-aided diagnostics. The precise identification and categorization of histopathological images are of paramount significance for early disease detection and treatment. In the diagnostic process of pathologists, a multi-tiered approach is typically employed to assess abnormalities in cell regions at different magnifications. However, feature extraction is often performed at a single granularity, overlooking the multi-granular characteristics of cells. To address this issue, we propose the Fuzzy-guided Multi-granularity Deep Neural Network (FMDNN). Inspired by the multi-granular diagnostic approach of pathologists, we perform feature extraction on cell structures at coarse, medium, and fine granularity, enabling the model to fully harness the information in histopathological images. We incorporate the theory of fuzzy logic to address the challenge of redundant key information arising during multi-granular feature extraction. Cell features are described from different perspectives using multiple fuzzy membership functions, which are fused to create universal fuzzy features. A fuzzy-guided cross-attention module guides universal fuzzy features toward multi-granular features. We propagate these features through an encoder to all patch tokens, aiming to achieve enhanced classification accuracy and robustness. In experiments on multiple public datasets, our model exhibits a significant improvement in accuracy over commonly used classification methods for histopathological image classification and shows commendable interpretability.
\end{abstract}


\section{Introduction}
\IEEEPARstart{H}{istopathological} images are of crucial significance in medical diagnosis and are essential for disease staging and treatment strategizing. The use of deep learning models for histopathological image classification tasks is a growing trend. Yang et al. \cite{ref1} realized a threshold-based tumor prioritization aggregation method for label inference within whole slide images and introduced a deep learning-based classifier tailored for lung lesion categorization, substantially advancing the identification of distinct lung cancer subtypes. Wang et al. \cite{ref2} used a fully convolutional neural network architecture in the extraction of pertinent deep features, to enable accurate predictions in lung cancer tissue histopathological image classification. Hence, deep neural networks have substantially improved the precision and efficiency of pathological diagnosis, thereby facilitating early intervention and treatment.

Histopathological images commonly encompass a variety of morphological features, including cell size, shape, and color, as well as the size and shape of cell nuclei and the distribution of chromatin. These features exhibit alterations indicative of pathological changes in cells or tissues, underscoring the significance of accurate feature extraction. Widely used approaches for feature extraction include Convolutional Neural Networks (CNNs) and Attention Mechanisms \cite{ref3}. The CNN operates by sliding convolutional kernels over an image, performing nonlinear transformations on the covered regions and extracting features at various hierarchical levels. Wahab et al. \cite{ref4} devised a Hybrid-CNN model by integrating two-stage CNNs through weight transfer and custom layers, which accomplished the accurate classification of mitotic and non-mitotic cells. Attention mechanisms have been extensively utilized in histopathological image classification tasks. Differing from CNNs, attention mechanisms capture global dependencies by focusing on all elements of the input sequence, employing weighted aggregations of these features to construct higher-level representations. Sadafi et al. \cite{ref5} applied an attention mechanism to the classification of hereditary blood disorders, showcasing its efficacy in enhancing a model's focus on pathological cell samples and increasing classification accuracy. Valanarasu et al. \cite{ref6} introduced a gated axial-attention model, incorporating additional control mechanisms within the self-attention module. The model's global branch captures long-range dependencies to learn contextual features across the entire image, while the local branch refines finer features through patch-level operations, yielding notable results in medical image analysis.

The above methods neglect the features of histopathological images at diverse granularity levels, limiting feature extraction to a singular scale, which leads to the incomplete capture of intrinsic feature information among cells. To address this, Li et al. \cite{ref7} demonstrated the efficacy of embedding features of varying granularities during the extraction process, effectively mitigating the challenges of limited inter-class variance and substantial intra-class variance in pathological image analysis and leading to reduced sensitivity to image magnification. Similarly, Hashimoto et al. \cite{ref8} found through experiments that distinct class-specific features exist across various scales, achieving superior accuracy in tumor subtype classification, surpassing the performance of expert pathologists. We can conclude from the above findings that to employ multi-granularity theory in histopathological image classification tasks can enable the extraction of medical pathological features at diverse granularity levels, leading to significant improvement in classification accuracy.

The fusion of multi-granular features may encounter information redundancy. Sinha et al. \cite{ref9} employed a multi-scale guided attention network in medical image analysis, leveraging additional losses between modules to disregard irrelevant information and accentuate relevant feature correlations, focusing on more discriminative regions. Xue et al. \cite{ref10} proposed a deep neural network with a global guidance block. Utilizing multi-level integrated feature maps as guiding information, they employed spatial and channel domain distant non-local dependencies. Their approach outperformed other medical image techniques, particularly in breast ultrasound lesion detection.

We incorporate the theory of fuzzy sets to guide the operation, where membership functions characterize the degree of membership of elements, which allows for the expression of the uncertainty and ambiguity of pixels in medical images\cite{ref11}. To capture the overlapping regions and indistinct boundaries in images and to mitigate potential disturbances caused by noise, artifacts, and lighting variations, membership functions allow each pixel to flexibly express its membership degree to different tissues and structures. Ding et al. \cite{ref12} used Interval Type-2 Fuzzy Clustering and metaheuristics to enhance the objective function of conventional fuzzy c-means clustering, incorporating spatial information based on neighboring local windows of superpixels, achieving effective segmentation of radiographic images. Ahmed et al. \cite{ref13} combined the Mamdani fuzzy model with adaptive neuro-fuzzy inference systems in a chronic kidney disease diagnostic system. It can be seen that the adoption of fuzzy set theory in medical imaging offers substantial support for clinical diagnosis and treatment \cite{ref14}.

We propose FMDNN, which is depicted in Fig. 1. When conducting feature extraction on histopathological images, we consider the medical structural properties of cells and individually extract tissue features at coarse, medium, and fine granularities. A fuzzy-guided cross-attention module incorporates universal fuzzy features into patch tokens, thus guiding features from different granularity. The main contributions of this paper are as follows:
\begin{enumerate}
    \item We introduce a multi-granular feature extraction method and employ feature visualization to investigate the semantic scales of features at different granularities, which contributes to a more comprehensive interpretation of information embedded in histopathological images;
    \item We use fuzzy logic to amalgamate key features from images through the computation of universal fuzzy features using three membership functions, which are utilized to globally guide the learning of multi-granular features. Experiments demonstrate the effectiveness of this approach in enhancing classification accuracy during model-guided training while minimizing interference from unrelated features;
    \item To address information redundancy in multi-granular feature extraction, we employ a fuzzy-guided approach to enhance the importance of key features. Continuously guiding features at three granularities with universal fuzzy features reduces interference from irrelevant information, significantly improving classification performance.
\end{enumerate}

The remainder of this article is structured as follows. Section II presents an overview of relevant methods and the utilization of fuzzy logic in classification. Section III introduces the three modules of our model. Sections IV and V describe comparative experiments and ablation studies that validate the effectiveness of the proposed model. In section VI, we summarize our work and propose future research directions.

\section{Related Work}
\subsection{Existing Classification Methods}
CNNs and attention mechanisms have shown notable success in image classification. Current methods can be broadly classified as based on (1) multiple feature inputs; (2) attention mechanisms; or (3) other functional modules.

Among methods based on multiple input features \cite{ref15}, Wang et al. \cite{ref16} proposed a pyramid vision Transformer model that can generate feature pyramids akin to CNNs, thus achieving multiscale feature integration. Gradually reducing image spatial resolution, Zheng et al. \cite{ref17} employed a Transformer framework to serially process images, thereby realizing a pure self-attention feature representation encoder. Tang et al. \cite{ref20} introduced adaptive convolution with a globally complementary context for multiscale design, allowing for the comprehensive acquisition of multimodal information from various scales.

Among methods using diverse attention mechanisms, Yuan et al. \cite{ref22} used an attention mechanism, Vision Outlooker, effectively encoding fine-grained features into Vision Transformer (ViT) token representations, thereby enhancing classification performance. Chu et al. \cite{ref23} introduced the Twins model, featuring spatially separable self-attention, which optimizes and accelerates the computational process of deep learning models primarily through matrix multiplication.

Among methods reliant on other functional modules, Touvron et al. \cite{ref25} introduced a teacher-student distillation training strategy tailored for ViT, which incorporated token-based distillation, enabling state-of-the-art (SOTA) results using ImageNet without external data. Chen et al. \cite{ref26} used memory-driven Transformers to generate medical reports, with memory-driven modules to capture essential generated information.

\subsection{Fuzzy logic}
Proposed by Zadeh in 1965, Fuzzy Set Theory addresses problems involving uncertainty and fuzziness \cite{ref27} and can model and represent image features. Unlike traditional Boolean logic, an element can belong to multiple sets, with an assigned membership degree between 0 and 1. The membership degree of element x in fuzzy set A can be expressed as $\mu A(x)\in[0,1]$. The flexibility of fuzzy logic is well suited to problems with fuzziness and uncertainty.

The combination of fuzzy sets with images is typically achieved through either: (1) fuzzy algorithms primarily based on machine learning; or (2) fuzzy feature processing predominantly based on deep neural networks. Wan et al. \cite{ref28} introduced the sparse fuzzy two-dimensional discriminant local preserving projection algorithm to mitigate the impact of variations, overlaps, and sparse points in images. Bhalla et al. \cite{ref29} proposed a hybrid fuzzy CNN approach, using fuzzy sets to eliminate uncertainty in images.

The complexity of configuring fuzzy membership functions requires task-specific adjustments, which significantly impacts model performance. The application of fuzzy set theory in the medical image domain is the subject of ongoing research \cite{ref35}.

\begin{figure*}[b]
\centering
\includegraphics[width=6.5in]{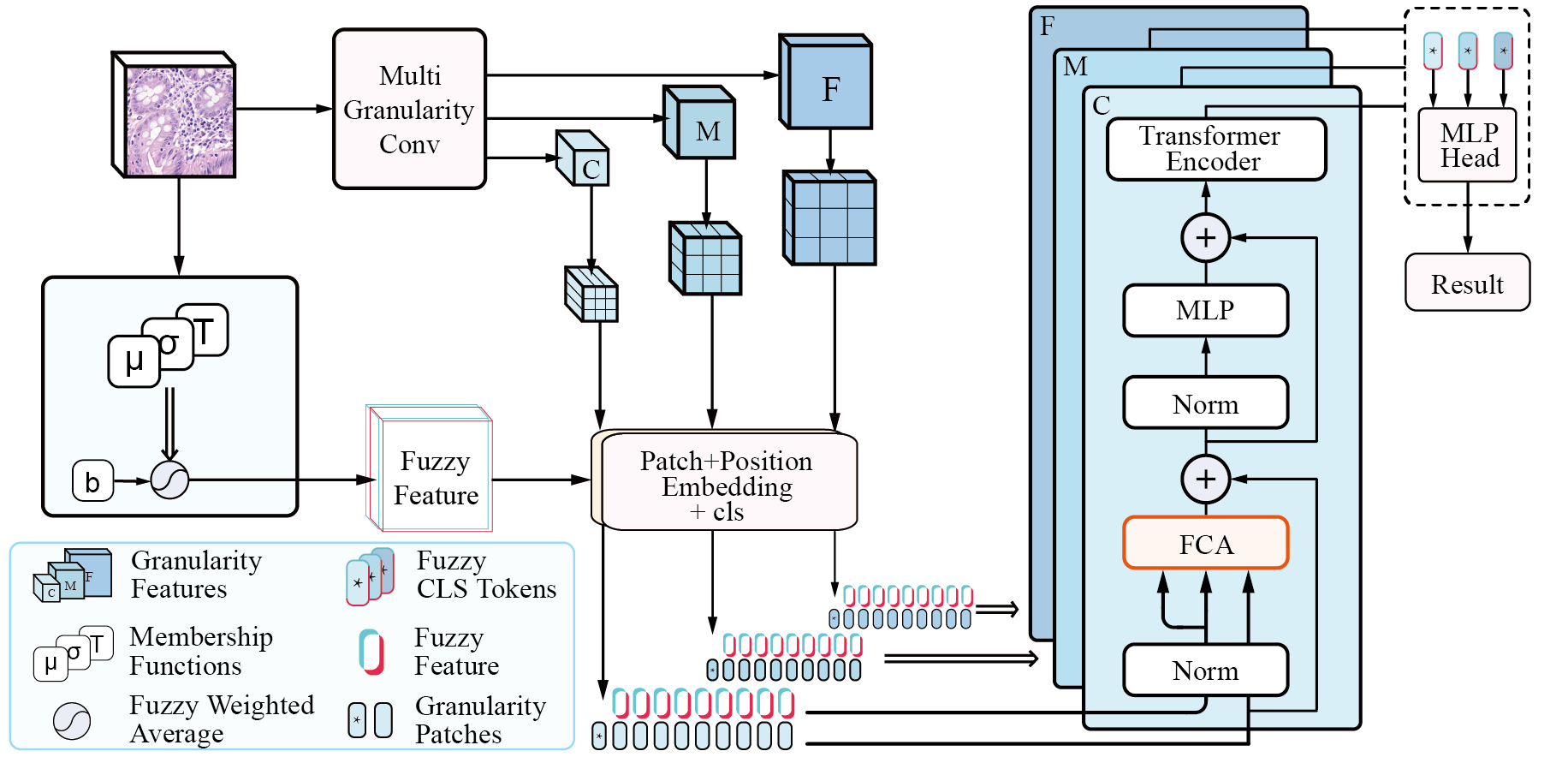}
\caption{Overview of proposed FMDNN.}
\label{fig_sim}
\end{figure*}

\subsection{Fuzzy Deep Neural Network}
Fuzzy logic is more adaptable in processing medical images \cite{ref30}, which often contain structures and pathological changes with fuzzy boundaries. Ding et al. \cite{ref31} introduced FTransCNN, a composite model with a fuzzy fusion module that amalgamates features extracted from CNN and Transformer architectures. Hu et al. \cite{ref32} addressed challenges in brain imaging, such as varying noise levels, indistinct boundaries, and artifacts, by combining improved fuzzy clustering with the HPU-Net model for brain image processing and disease diagnosis prediction.

Fuzzy set methods have also been used in the fusion of multimodal medical images \cite{ref36}. By defining multiple membership functions and fuzzy rules, they integrate information from various modalities, thereby enhancing the accuracy of classification and diagnosis. Wang et al. \cite{ref33} fused multiple CNNs and fuzzy neural network Gabor representation, filtering CT and MR image sets to realize distinct feature representations. Fuzzy set methods have been applied to medical image classification and diagnosis tasks. Das et al. \cite{ref34} introduced a feature-driven linguistic fuzzy neural model for disease classification analysis in medical data, leveraging linguistic fuzzification to generate membership values to handle uncertainty and integrating feature extraction algorithms within the fuzzy neural architecture to extract crucial features in medical data analysis.

Despite successes in medical image classification, several issues in histopathological images have been overlooked. Specifically, there is a lack of effective utilization of multi-granularity, as well as a lack of appropriate methods for guiding the learning of multi-granular features. A multi-granular feature extraction module is introduced that facilitates the extraction of key features across diverse granularity levels. We use fuzzy logic to compute and integrate universal features from the image. Fuzzy-guided cross-attention is introduced for feature fusion. Throughout the training process, universal fuzzy features continuously guide multi-granular features, overcoming information redundancy in multi-granularity learning. The model can acquire more effective classification features.

\section{Proposed Method}
As shown in Fig. 1, FMDNN conducts feature extraction on the input image at three distinct granularities, whose features are concatenated with the universal fuzzy feature and fed into the corresponding encoder, yielding classification tokens for each granularity. The final classification results are derived through linear transformation for feature fusion and dimension alignment.

\begin{figure}[!t]
\centering
\includegraphics[width=3in]{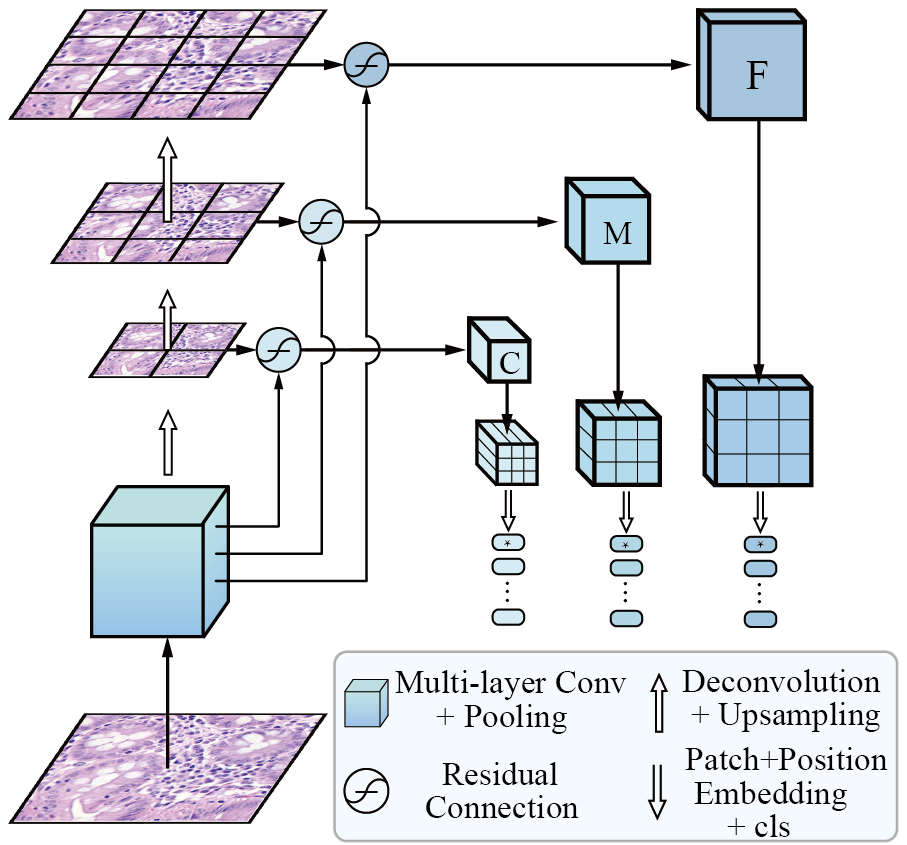}
\caption{Image feature extraction process for obtaining fine-, medium-, and coarse-grained features.}
\label{pic2}
\end{figure}

\subsection{Multi-granular Feature Extraction Module}
Relying solely on singular features has proven inadequate in feature extraction from medical images \cite{ref37,ref38,ref39}. Given the unique characteristics of histopathological images, pathologists typically employ multi-granularity when classifying them. This involves observing the overall tissue morphology at low magnification to identify obvious abnormal regions. At medium magnification, attention shifts to cell arrangement, nucleus morphology, and cell relationships. High magnification is then used to examine features such as nuclear characteristics, intracellular organelles, and clarity of cell boundaries. Integrating observations across multiple microscopic levels provides a more comprehensive understanding of lesion nature and severity \cite{ref40}.

Inspired by the multiscale diagnostic approach employed in histopathological image classification, we incorporate CNN-based multi-granular feature extraction to extract features separately on fine-grained $x_{Fine}\in\mathbb{R}^{\frac{H}{2r}\times{\frac{W}{2r}}\times C_{Fine}}$, medium-grained $x_{Medium}\in\mathbb{R}^{\frac{H}{r}\times{\frac{W}{r}}\times C_{Medium}}$, and coarse-grained $x_{Coarse}\in\mathbb{R}^{H\times W\times C_{Coarse}}$ levels. As illustrated in Fig. 2, an encoder-decoder structure is employed, and multiple layers of convolution and pooling are used to extract image features and reduce granularity. In the decoder, deconvolution and upsampling are used while extracting the desired granularity of features. The design ensures that the dimensions of the three granularity blocks remain the same during block embedding, reducing fusion costs and accelerating computation. As demonstrated by Kaul et al. \cite{ref43}, skip connections within a network facilitate the smooth propagation of gradients across the entire network, effectively preserving feature maps, particularly in capturing intricate details within fine-grained levels. The final representation of the features is $x=(x_{Fine},x_{Medium},x_{Coarse})$, where $x_{Fine}$, $x_{Medium}$, and $x_{Coarse}$, respectively, represent fine-, medium-, and coarse-grained features.

Multi-granular features $x_{Fine}$, $x_{Medium}$, and $x_{Coarse}$ are reshaped into flattened 2D feature blocks $x_{FineP}\in\mathbb{R}^{N\times((P/2r)^{2}\cdot C_{Fine})}$, $x_{MediumP}\in\mathbb{R}^{N\times((P/r)^{2}\cdot C_{Medium})}$, and $x_{CoarseP}\in\mathbb{R}^{N\times((P^{2}\cdot C_{Coarse})}$, respectively. Here, $(H,W)$ represents the original resolution of the features, $C$ is the number of channels, $(P,P)$ is the resolution of each feature block, and $N$ is the number of generated feature blocks, which is also the effective input sequence length for fusion with fuzzy features. Our design ensures that the dimension of $N$ matches across granularities, significantly reducing computational complexity during fusion with fuzzy features. Similar to the $\textbf{[class]}$ token in ViT, we add learnable embeddings to each embedded feature block sequence, and the final output can be represented as:
\begin{equation}
    z_{0}^{F}=[x_{\mathrm{class}}^{F};x_{P}^{1F}E^{F};x_{P}^{2F}E^{F};\cdots;x_{P}^{NF}E^{F}]+E_{pos}^{F},
\end{equation}
\begin{equation}
    z_{0}^{M}=[x_{\mathrm{class}}^{M};x_{P}^{1M}E^{M};x_{P}^{2M}E^{M};\cdots;x_{P}^{NM}E^{M}]+E_{pos}^{M},
\end{equation}
\begin{equation}  z_{0}^{C}=\bigl[x_{\mathrm{class}}^{C};x_{P}^{1C}E^{C};x_{P}^{2C}E^{C};\cdots;x_{P}^{NC}E^{C}\bigr]+E_{pos}^{C},
\end{equation}
where variables $E_{pos}^{F}\in\mathbb{R}^{(N+1)\times D}$, $E_{pos}^{M}\in\mathbb{R}^{(N+1)\times D}$, and $E_{pos}^{C}\in\mathbb{R}^{(N+1)\times D}$ are introduced, and $E^{F}\in\mathbb{R}^{\left(\left(\frac{p}{2r}\right)^{2}\cdot C_{F}\right)\times D}$, $E^{M}\in\mathbb{R}^{\left(\left(\frac pr\right)^{2}\cdot C_{M}\right)\times D}$, and $E^{C}\in\mathbb{R}^{((P^{2}\cdot C_{C})\times D}$. Similar to ViT, our model adopts the position embedding strategy when dealing with image data, embedding position information in each feature block to retain the accurate position of each pixel.

\subsection{Universal Fuzzy Feature Module}
We adopt fuzzy set theory to address information redundancy in multi-granular feature extraction, treating image pixels as elements and describing their membership to fuzzy concepts by membership functions. This approach can capture uncertainty and fuzziness in images. By setting different membership functions and thresholds, it is possible to accurately extract common features of malignant cells, thus providing more precise information for image classification.

In the Universal Fuzzy Feature Algorithm, an image $I$ is converted to grayscale and normalized to the range $[0,1]$. When extracting the fuzzy features of the image, each pixel $I_{x}$ is considered a fuzzy set, and different membership functions are applied to each image. We extract three fuzzy features. The obtained fuzzy universal feature set is $\{I_{\mu},I_{\sigma},I_T\}$, whose definition depends on the membership functions, which describe the degree to which each pixel belongs to a certain fuzzy concept.

The universal fuzzy feature $z^{Fuzzy}$ is obtained through fuzzy operations. During the model learning process, the membership functions can help it to better understand the meaning of each pixel, including its potential membership categories and degrees.

To extract multiple features from the image, we select multiple membership functions, each corresponding to a specific feature representation method. We choose Gaussian, Sigmoid, and Trapezoidal functions, enabling more effective guidance of the model in learning key features through the universal fuzzy features, which are used to construct fuzzy sets. Through fuzzy set intersection operations, we extract the universal features of the image. The first membership function is defined using the Gaussian function, calculating the membership degree of each pixel by applying the Gaussian function to its grayscale value,
\begin{equation}\mu_{x}=\frac{1}{\sigma\sqrt{2\pi}}\exp\left(-\frac{(x-\mu)^{2}}{2\sigma^{2}}\right),\end{equation}
where $\mu$ is the mean, $\sigma$ is the standard deviation, $x$ is the grayscale value of the pixel, and $\mu_x$ is its membership degree. Next, we define the Sigmoid membership function:
\begin{equation}\sigma_{x}=\frac{1}{1+\exp\left(-\alpha(l_{x}-\beta)\right)},\end{equation}
where $I_x$ is the brightness value of pixel $x_i$, and $\alpha$ and $\beta$ are parameters of the Sigmoid function, used to adjust its shape and position, respectively.

Finally, we define the Trapezoidal membership function

\begin{equation}
    T(x;a,b,c,d)=\begin{cases}0,&x\le a\\ \frac{x-a}{b-a},&a<x\le b\\ 1,&b<x\le c\\ \frac{d-x}{d-c},&c<x<d\\ 0,&x\ge d,\end{cases}
\end{equation}
where a and b are the respective left and right endpoints; and c and d are the ascending and descending inflection points, respectively. The function is controlled by ascending and descending slopes at its respective left and right ends, while the middle part has a membership degree of 1.

We use fuzzy set theory to effectively fuse features extracted from various membership functions, forming a more comprehensive and accurate universal feature for the image. We introduce a fuzzy weighting strategy to fuse uncertain data, aiming to integrate the three features for a more comprehensive feature description. We assign weights $w_\mu$, $w_\sigma$, and $w_T$ to fuzzy features $I_{\mu}$, $I_{\sigma}$, and $I_T$, respectively, representing their importance in the overall description, and satisfying the condition $w_\mu + w_\sigma + w_T = 1$. We introduce a bias term B to adjust the baseline level of fuzzy feature fusion and enhance the model's flexibility.

The fused feature, i.e., the universal fuzzy feature obtained from the above variables and parameters, can be calculated as
\begin{equation}z^{Fuzzy}=w_{\mu}\times I_{\mu}+w_{\sigma}\times I_{\sigma}+w_{T}\times I_{T}+b.\end{equation}

In the process of fuzzy fusion, the membership degree of each data point determines its contribution to the mean value. Through this method, we can extract rich fuzzy features that can be used in subsequent deep learning models.

\subsection{Fuzzy-guided Cross-attention Module}

\begin{figure*}[!t]
\centering
\includegraphics{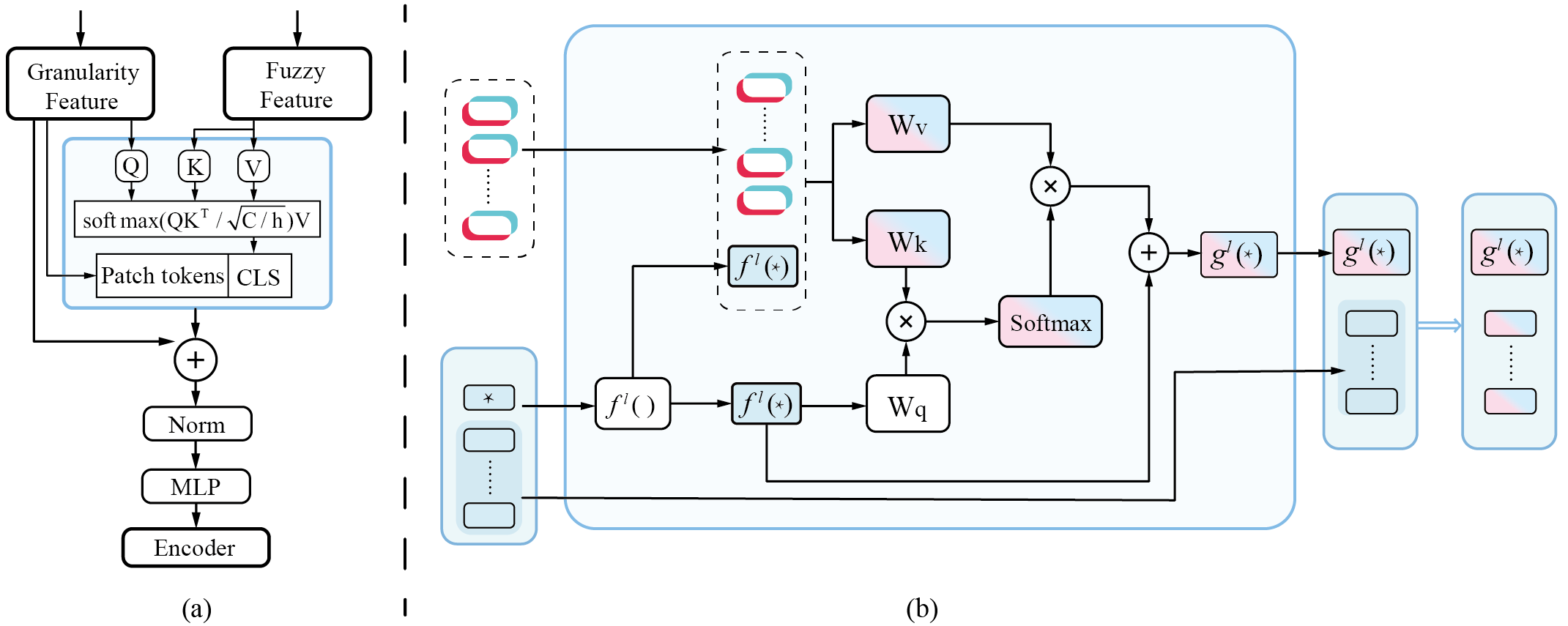}
\caption{FCA Framework: (a) FCA operator; (b) visualization of workflow.}
\label{pic3}
\end{figure*}

Weighted averaging is commonly used in fusion approaches, where the fuzzy feature matrix $x^{Fuzzy}$ is normalized and directly added to a Transformer encoder or another fusion module \cite{ref44}. However, this method assigns the same weight to all models. Using the fused output as input can enhance model expressiveness and generalization. However, in multi-granularity models, this approach may lead to information redundancy, or even the loss of critical features in certain granularities. Zhang et al. \cite{ref45} introduced an Edge-Guided module for medical image datasets, which learns edge-aware representations and preserves local edge features at early encoding layers, with good results.

We propose fuzzy-guided cross-attention (FCA) for fusion, where universal fuzzy features, serving as the driving information, are deeply integrated with features of different granularities through cross-attention, rather than being fused using addition. As shown in Fig. 3(a), the key characteristic of this process is the ability to continuously drive pixel-level information using the universal fuzzy feature, preventing overfitting when continuously learning on three individual granularities. After interacting with the fuzzy patch tokens, the guiding information from the classifier is spread to the patch tokens through the Transformer encoder, continuously guiding the learning of multi-granular features. This is expressed as (\ref{equ1}), where $\ell=1,2,\ldots , L$.
\begin{equation}
\label{equ1}
    \begin{split}
        z^{\prime}{}_{\ell}^{G}=\mathrm{FCA}\left(\mathrm{LN}(z^{Fuzzy}),\mathrm{LN}\big(z_{\ell-1}^{G}\big)\right)+z_{\ell-1}^{G},
    \end{split}
\end{equation}

Specifically, for different granularity branches $z_l^G$, we have $z_{\ell}^{G}=[f^{l}(z_{cls}^{G})\|z_{patch}^{Fuzzy}]$, where $G\in(C,M,F)$ represents the three granularities, $f^{l}(\cdot)$ is the projection function used for dimension alignment, and $\mathrm{LN(\cdot)}$ is the layer normalization function. After using the guiding information for fusion, it will re-enter the next Transformer encoder and interact with its corresponding patch tokens:
\begin{equation}z^{\prime\prime}{}_{\ell}^{G}=\mathrm{MLP}\big(\mathrm{LN}\big(z{'}_{\ell}^{G}\big)+z{'}_{\ell}^{G}\big),\ell=1,2,\ldots ,L,\end{equation}
\begin{equation}z_{\ell}^{G}=\mathrm{Encoder}\bigl(z^{\prime\prime}{}_{\ell}^{G}\bigr),\ell=1,2,\ldots ,L,\end{equation}
which promotes the propagation of guiding information and enriches the representation of each image patch, leading to improved overall model accuracy and robustness. Finally, by aggregating the representations processed through multiple Transformer encoders, we obtain the final classification results:
\begin{equation}y=\mathrm{LN}\Big(([z_{L}^{F}]_{cls},[z_{L}^{M}]_{cls},[z_{L}^{C}]_{cls})\cdot W_{x}\Big).\end{equation}

It is worth noting that we do not use traditional cross-attention, i.e., we do not simply change the sources of $Q$, $K$, and $V$, which, as shown through experiments in Section IV, does not lead to significant improvements. Fig. 3(b) shows the algorithm, where fusion involves classification and patch tokens from multiple granularity branches, as well as fuzzy patch tokens. Similar to CrossViT \cite{ref24}, we use fuzzy patch tokens from the fuzzy features as guiding information to interact with multi-granularity patch tokens, enabling deep guidance.

FCA is implemented as
\begin{equation}Q=z^{Fuzzy}W_{q},K=z_{\ell}^{G}W_{k},V=z_{\ell}^{G}W_{\nu}\end{equation}
\begin{equation}\mathrm{FCA}(Q,K,V)=\mathrm{softmax}\left(\frac{\mathrm{QK}^{T}}{\sqrt{C/h}}\right)V,\end{equation}
where $W_{q},W_{k},W_{v}\in\mathbb{R}^{C\times\left(\frac{C}{h}\right)}$ are learnable parameters; and $C$ and $h$ are the embedding dimension and number of attention heads, respectively.

It has been demonstrated that the multi-granularity branches primarily serve the role of feature extraction, with the fuzzy feature branch contributing supplementary information \cite{ref24}. Consequently, a lightweight fuzzy feature branch alone is sufficient to effectively guide the multi-granularity branches. We apply cross-attention to fuse multi-granular features, as the main information sources, with universal fuzzy features, which are treated as supplementary information. FMDNN achieves higher accuracy than other fusion methods.

\subsection{Model framework}
Algorithm 1 shows the implementation of the model. An initial step employs a multi-granularity convolution technique to extract features from the original image at various granular levels. Through the fusion of membership functions tailored to distinct feature points, universal fuzzy features of the original image are extracted. Extracted multi-granular features and universal fuzzy features are separately embedded with patches at different granularities, harmonizing dimensions and reducing subsequent algorithmic complexity.

Next is the fuzzy-guided cross-attention module, which utilizes the universal fuzzy features to guide training at distinct granularities. By employing a multi-granularity classifier as an intermediary, features from different granularities and universal fuzzy features are integrated through fusion learning of feature information and are subsequently back-projected into their respective branches.
Each training instance comprises a granularity-specific feature and a fuzzy feature. During each training iteration, the multi-granularity classification token interacts with a fuzzy patch token that serves as guiding information. Post-fusion, with this guiding information, the multi-granularity classification token engages once again with its corresponding patch token within the subsequent Transformer encoder, enabling the transfer of acquired guiding information to the associated patch token, enriching the representation of each image block.

\begin{small}

\begin{algorithm}[ht]
\SetAlgoLined
\LinesNumbered
\SetNlSty{}{}{:}
            \caption{FMDNN}
                \KwIn{image $I$ to be extracted; membership functions $\mu_{x}(\cdot)$, $\sigma_{x}(\cdot)$, and $T_{x}(\cdot)$; shape and position adjustment parameters $\alpha$ and $\beta$; rise and fall slope a, b, c, d.}
            \KwOut{classification MLP head $y$}
\BlankLine
            \For{each image $I$}{ /* Calculate the membership degree for each pixel */
            
            $\mathbf{z}^{C},\mathbf{z}^{M},\mathbf{z}^{F}\gets [\mathbf{x}_{\mathrm{class}}^{G};\mathbf{x}_{P}^{1G}E;\mathbf{x}_{P}^{2G}E;\cdots;\mathbf{x}_{P}^{NG}E]+E_{pos}^{G}$
            
                \For{each pixel in $I$}{
                $\mu={\frac{1}{n}}\sum_{t-1}^{n}x_{i}$

                $\sigma=\sqrt{\frac{1}{n}\sum_{i=1}^{n}(x_{i}-\mu)^{2}}$

                /* Define fuzzy membership functions */

                $\mathbf{I}_{\mu}\gets\mu_{x}(I,\mu,\sigma)$

                $\mathbf{I}_{\sigma}\gets\sigma_{x}(I,\alpha,\beta)$

                $\mathbf{I}_{T}\gets T(I;a,b,c,d)$
                }
            Set $\mathbf{I}\gets(\mathbf{I}_{\mu},\mathbf{I}_{\sigma},\mathbf{I}_{T},E)$ /* Get fuzzy feature matrix */

            Set $\mathbf{W}\gets(\mathbf{w}_{\mu},\mathbf{w}_{\sigma},\mathbf{w}_{T},b)^{T}$ /* Obtain universal fuzzy features through fuzzy calculation */

            $\mathbf{z}^{Fuzzy}\gets\mathbf{I}\cdot\mathbf{W}$

            $[\mathbf{z}_{L}^{C}]_{cls}\gets\mathbf{Encoder}\big(\mathbf{FCA}(\mathbf{z}^{Fuzzy},\mathbf{z}^{C})\big)$

            $[\mathbf{z}_{L}^{M}]_{cls}\gets\mathbf{Encoder}\big(\mathbf{FCA}(\mathbf{z}^{Fuzzy},\mathbf{z}^{M})\big)$

            $[\mathbf{z}_{L}^{F}]_{cls}\gets\mathbf{Encoder}\big(\mathbf{FCA}(\mathbf{z}^{Fuzzy},\mathbf{z}^{F})\big)$

            $y\gets\big(([\mathbf{z}_{L}^{F}]_{cls},[\mathbf{z}_{L}^{M}]_{cls}[\mathbf{z}_{L}^{C}]_{cls})\cdot \mathbf{W}_{x}\big)$
            }
            
            \textbf{return} $y$
\end{algorithm}
\end{small}

\section{Experiments}

\subsection{Datasets and Configurations}
We evaluated the performance of FMDNN in medical image classification on five publicly available pathological image datasets, as shown in Table I. These include diverse tissue types, varying numbers of categories, and various granularity attributes. These are as follows:
\begin{enumerate}
    \item   Lung and Colon Cancer Histopathological Images    (\textbf{LC})
    \item   NCT-CRC-HE-100K  (\textbf{NCT})
    \item   APTOS 2019 Blindness Detection  (\textbf{Bl})
    \item   HAM10000  (\textbf{HAM})
    \item   Kvasir  (\textbf{Kv})
\end{enumerate}

Within the model, we configured different granularity blocks as 224 × 224 × 3, 112 × 112 × 12, and 56 × 56 × 48. We divided the dataset into training, validation, and testing sets in a 70:15:15 ratio. Data augmentation techniques such as random cropping, horizontal flipping, and rotation were employed to enhance the model's generalization capabilities. We utilized cross-entropy as the loss function and trained the model using stochastic gradient descent with a batch size of 64. The initial learning rate was set to 0.001, with a decay factor of 0.01.

\begin{table}[t] 
\caption{Number and Category of Datasets}
\label{tab:dataset}
\begin{tabular}{lccccc}
\toprule

\textbf{Dataset}  & \textbf{LC}  & \textbf{NCT}   & \textbf{Bl}   & \textbf{HAM}  & \textbf{Kv}                                                    \\ \hline
\begin{tabular}[c]{@{}l@{}}Num of\\ Images\end{tabular}   & 25000   & 100000  & 5590    & 10015   & 8000                                                  \\
\begin{tabular}[c]{@{}l@{}}Num of\\ Classes\end{tabular}     & 5     & 9     & 5      & 7   & 8                                                     \\
\begin{tabular}[c]{@{}l@{}}Multi-\\ granularity\\ attributes\end{tabular} & Obvious & Obvious & \begin{tabular}[c]{@{}c@{}}Not\\ Obvious\end{tabular} & \begin{tabular}[c]{@{}c@{}}Not\\ Obvious\end{tabular} & \begin{tabular}[c]{@{}c@{}}Not\\ Obvious\end{tabular} \\ 
\bottomrule
\end{tabular}
\end{table}

\subsection{Experimental Setup}
We compared FMDNN against several classification models on different datasets:
\begin{enumerate}
    \item Resnet50 leverages residual connections to facilitate optimization and mitigates gradient vanishing. We utilized pretrained weights on the ImageNet dataset. We denote this as \textbf{Resnet50\_pre} \cite{ref51};
    \item We used the ViT-base model with pretrained weights from the ImageNet-21k dataset, denoted as \textbf{ViT-base\_pre} \cite{ref52};
    \item With medical images, HiFuse adopts a three-branch hierarchical multi-scale feature fusion network structure, with the advantages of Transformer and CNN across multiple scale levels. We denote this as \textbf{HiFuse} \cite{ref53};
    \item MLP-Mixer uses an architecture based solely on multi-layer perceptrons. We denote this as \textbf{MLP-Mixer} \cite{ref54};
    \item RSP+CR uses a semi-supervised learning framework incorporating knowledge distillation. It extracts context information at various resolutions within the multi-layer pyramid structure present in histopathological images. We denote this as \textbf{RSP+CR} \cite{ref55}.
\end{enumerate}

We employed accuracy (ACC), true-positive rate (TPR), true-negative rate (TNR), positive predictive value (PPV), and F1-score (F1) as evaluation metrics.

\subsection{Experimental Results}
Table II shows the results of our comparative experiments.
FMDNN performed notably on LC and NCT, with a 99.2\% ACC and 98.9\% F1-score on LC, and values of 98.1\% on both metrics on NCT.

\begin{table}[t]
    \centering
    \caption{Results of Comparative Experiments}
    \label{tab:performance_metrics}
    
    \begin{tabular}{p{0.5cm}p{1.6cm}p{0.5cm}p{0.5cm}p{0.5cm}p{0.5cm}p{0.5cm}}
        \toprule
         \multicolumn{1}{c}{\textbf{Dataset}} &  \multicolumn{1}{c}{\textbf{Model}} & \textbf{ACC (\%)} & \textbf{TPR (\%)} & \textbf{TNR (\%)} & \textbf{PPV (\%)} & \textbf{F1 (\%)} \\
        \midrule
        LC & Resnet50\_pre & 96.5 & 96.4 & 95.8 & 96.1 & 96.2 \\
        & ViT-base\_pre & 94.8 & 95.2 & 93.6 & 94.2 & 94.7 \\
        & HiFuse & 97.6 & 97.0 & 97.9 & 98.5 & 98.0 \\
        & MLP-Mixer & 93.1 & 96.3 & 93.2 & 95.2 & 95.4 \\
        & RSP+CR & 98.5 & 94.3 & 95.7 & 95.2 & 94.1 \\
        & \textbf{FMDNN} & \textbf{99.2} & \textbf{98.7} & \textbf{99.6} & \textbf{99.4} & \textbf{98.9} \\
        \midrule
        NCT & Resnet50\_pre & 94.6 & 95.2 & 93.1 & 93.5 & 93.8 \\
        & ViT-base\_pre & 90.8 & 90.6 & 89.9 & 90.2 & 90.4 \\
        & HiFuse & 96.2 & 95.7 & 94.7 & 96.9 & 96.6 \\
        & MLP-Mixer & 95.6 & 96.3 & 98.0 & 97.6 & 96.6 \\
        & RSP+CR & 97.6 & 93.8 & 95.5 & 95.4 & 93.9 \\
        & \textbf{FMDNN} & \textbf{98.1} & \textbf{98.0} & \textbf{98.6} & \textbf{98.2} & \textbf{98.1} \\
        \midrule
        Bl & Resnet50\_pre & 75.4 & 70.5 & 73.2 & 72.1 & 71.5 \\
        & ViT-base\_pre & 69.3 & 36.5 & 49.4 & 46.7 & 43.1 \\
        & HiFuse & 81.3 & 58.6 & 63.4 & 68.9 & 65.4 \\
        & MLP-Mixer & 73.8 & \textbf{80.6} & 77.1 & 75.3 & 78.8 \\
        & RSP+CR & 80.7 & 75.7 & 71.0 & 78.6 & 74.8 \\
        & v & \textbf{88.2} & 78.5 & \textbf{81.1} & \textbf{80.6} & \textbf{79.2} \\
        \midrule
        HAM & Resnet50\_pre & 73.1 & 70.4 & 68.9 & 67.4 & 69.0 \\
        & ViT-base\_pre & 63.5 & 57.0 & 54.8 & 63.9 & 56.3 \\
        & HiFuse & 92.5 & 87.1 & 89.2 & 88.9 & \textbf{92.3} \\
        & MLP-Mixer & 57.2 & 59.8 & 57.9 & 56.2 & 59.1 \\
        & RSP+CR & 86.8 & 87.2 & 81.4 & 80.9 & 84.4 \\
        & \textbf{FMDNN} & \textbf{93.2} & \textbf{89.8} & \textbf{90.7} & \textbf{91.1} & 91.9 \\
        \midrule
        Kv & Resnet50\_pre & 72.6 & 76.7 & 79.0 & 74.2 & 73.6 \\
        & ViT-base\_pre & 76.5 & 75.8 & 74.8 & 75.9 & 76.2 \\
        & HiFuse & 84.8 & 85.1 & \textbf{90.1} & 84.3 & 85.6 \\
        & MLP-Mixer & 70.6 & 71.4 & 70.9 & 71.7 & 70.7 \\
        & RSP+CR & 80.2 & 75.7 & 79.9 & 76.3 & 81.8 \\
        & \textbf{FMDNN} & \textbf{90.4} & \textbf{89.7} & 89.0 & \textbf{90.2} & \textbf{88.6} \\
        \bottomrule
    \end{tabular}
\end{table}


MLP-Mixer and ViT-base\_pre had lower ACC values on LC, at 93.1\% and 94.8\%, respectively. This is attributed to the former's reliance on fully connected layers to capture image features and the latter's use of an attention mechanism to capture global contextual information. However, they neglect the uneven distribution of crucial features in images, leading to an excessive capture of features from healthy cells while overlooking critical features.

FMDNN achieved superior accuracy on Bl. Retinal cells are typically small and abundant and include intricate vascular networks compared with other cell tissues, posing a high demand on the model's generalization ability. HiFuse incorporates multi-scale (global-local) features and also performs well, emphasizing the crucial role of multi-scale attributes in medical image analysis, particularly tissue pathology image classification. FMDNN utilizes universal fuzzy features to annotate key features during classification, which mitigates the overlearning of irrelevant features. These benefit FMDNN in complex tasks such as retinal image classification, where features are not distinctly prominent.

 FMDNN showed good ACC, TPR, TNR, and PPV results on HAM, with an F1-score only 0.4\% less than second-best. On Kv, except for TNR, FMDNN achieved the highest scores across all evaluation metrics. FMDNN can deliver satisfactory results at tasks where multi-scale attributes are not prominent, as well as on datasets where multi-granularity attributes are not obvious. It also shows strong generalization performance.

The radar chart in Fig. 4 reveals that the proposed FMDNN continuously guides multi-scale features using universal fuzzy features, which enhances classification performance. In comparisons with different models on multiple datasets, FMDNN consistently demonstrated superior accuracy and robustness.


\begin{figure*}[!t]
\centering
\includegraphics[width=6in]{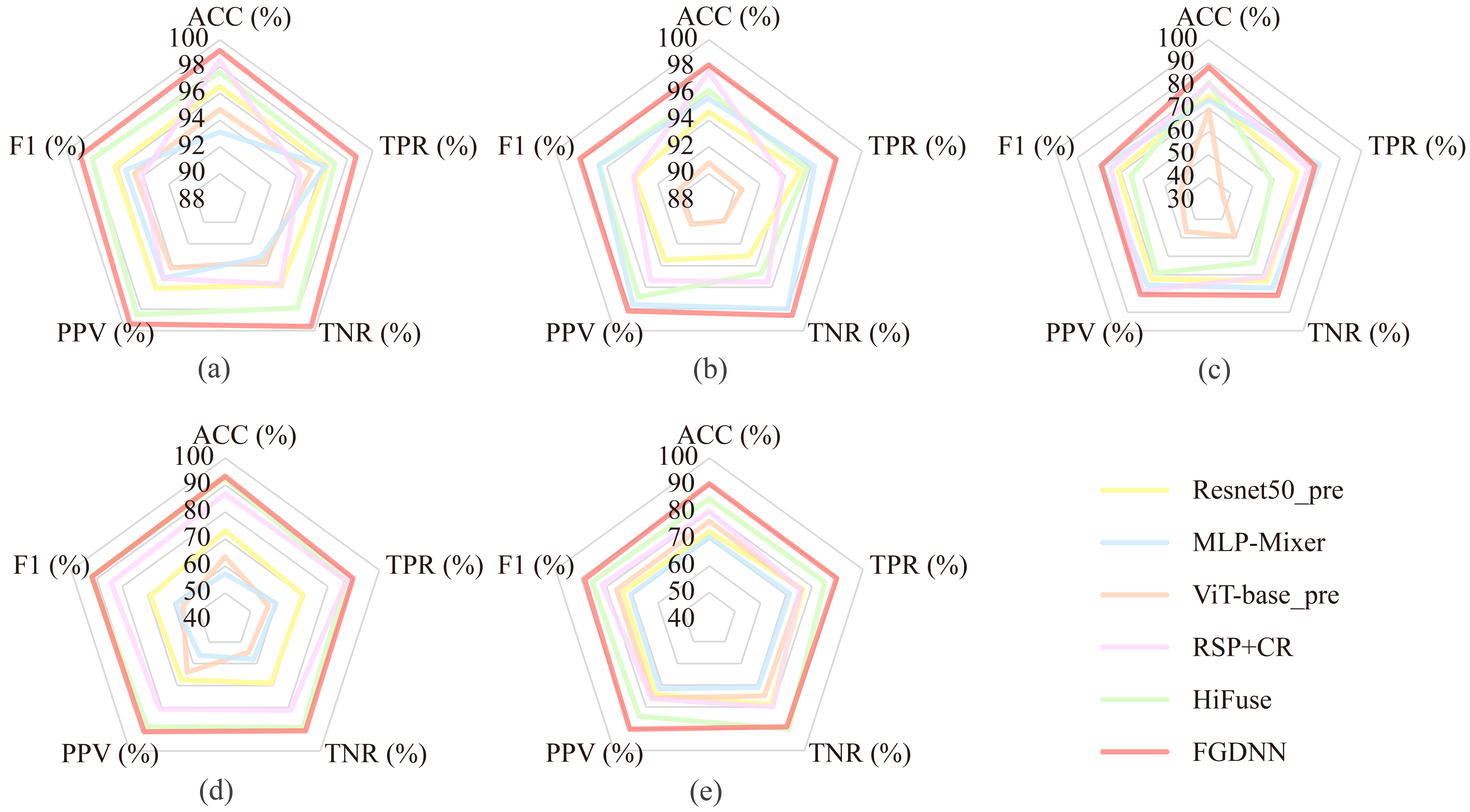}
\caption{Comparative experimental radar charts on different datasets: (a) LC; (b) NCT; (c) Bl; (d) HAM; (e) Kv.}
\label{pic4}
\end{figure*}

\section{Discussion}

\subsection{Comparison with State-of-the-Art Models}
We compared FMDNN with SOTA methods on various datasets, as shown in Table III.

FMDNN surpasses all SOTA models on LC and outperforms the majority of models on NCT, with an ACC only 0.5\% less than that of Srinidhi et al. It achieves a high F1-score of 98.1\%. Both of these datasets exhibit significant multi-granularity attributes, and FMDNN shows favorable results on other datasets as well. FMDNN outperforms all SOTA models on Bl, with an ACC of 88.2\%, which is significantly better than that of Kassani et al.'s model at 83.1\%. It also performs close to SOTA models on HAM and Kv.

FMDNN outperforms other classification algorithms on various datasets. However, its classification accuracy falls below expectations on HAM and Kv, mainly due to their unique characteristics, as they exhibit imbalanced numbers of samples across different categories, in which diseases share a high degree of visual similarity, posing a significant challenge to classification. In addition, the subtle nature of multi-granularity attributes in these datasets impacts the model's performance. However, FMDNN still achieves noteworthy results.

\begin{table}[h]
\renewcommand{\arraystretch}{1}
\caption{Comparison with SOTA Models}
\label{tab:performance}
\centering
\begin{tabular}{cccc}
\toprule
\textbf{Dataset} & \textbf{Author/Model} & \textbf{ACC (\%)} & \textbf{F1-score (\%)} \\
\hline
\multirow{3}{*}{LC} & Mehmood et al. \cite{ref56} & 98.4 & — \\
& Masud et al. \cite{ref57} & 98.9 & — \\
& \textbf{FMDNN} & \textbf{99.2} & \textbf{98.9} \\
\hline
\multirow{6}{*}{NCT} & Haghighi et al. \cite{ref58} & 93.1 & 92.7 \\
& Wang et al. \cite{ref59} & 96.5 & 94.8 \\
& Mormont et al. \cite{ref60} & 94.6 & 94.3 \\
& Srinidhi et al. \cite{ref55} & \textbf{98.6} & 93.4 \\
& Song et al. \cite{ref62} & 95.9 & 95.6 \\
& \textbf{FMDNN} & 98.1 & \textbf{98.1} \\
\hline
\multirow{3}{*}{Bl} & Mohanty et al. \cite{ref63} & 79.5 & — \\
& Kassani et al. \cite{ref64} & 83.1 & — \\
& \textbf{FMDNN} & \textbf{88.2} & \textbf{79.2} \\
\hline
\multirow{3}{*}{HAM} & Alenezi et al. \cite{ref65} & 95.7 & \textbf{93.4} \\
& Mehmood et al. \cite{ref66} & \textbf{96.9} & — \\
& \textbf{FMDNN} & 93.2 & 91.9 \\
\hline
\multirow{2}{*}{Kv} & Wang et al. \cite{ref67} & \textbf{94.8} & — \\
& \textbf{FMDNN} & 90.4 & \textbf{88.6} \\
\bottomrule
\end{tabular}
\end{table}



\begin{table}[t]
    \centering
    \caption{Results of Comparative Experiments}
    \label{tab:performance_metrics}
    
    \begin{tabular}{p{0.5cm}p{1.4cm}p{0.6cm}p{0.6cm}p{0.6cm}p{0.6cm}p{0.6cm}}
        \toprule
        \textbf{Dataset} & \multicolumn{1}{c}{\textbf{Model}} & \textbf{ACC (\%)} & \textbf{TPR (\%)} & \textbf{TNR (\%)} & \textbf{PPV (\%)} & \textbf{F1 (\%)} \\
        \midrule
        LC &     w/o MG & 94.2 & 93.6 & 93.8 & 92.7 & 92.9 \\
        &     w/o Fuzzy & 93.2 & 94.1 & 92.5 & 93.0 & 93.5 \\
        &     w/o C & 95.4 & 95.4 & 95.1 & 96.2 & 95.8 \\
        &     w/o M & 95.1 & 93.7 & 93.8 & 94.0 & 94.8 \\
        &     w/o F & 90.4 & 91.2 & 90.1 & 91.0 & 90.8 \\
        &     M-ADD & 89.4 & 88.4 & 89.9 & 88.4 & 88.2 \\
        &     M-CA & 96.1 & 96.2 & 95.3 & 95.8 & 95.9 \\
        &     \textbf{FMDNN} & \textbf{99.2} & \textbf{98.7} & \textbf{99.6} & \textbf{99.4} & \textbf{98.9} \\
        \midrule
        NCT &     w/o MG & 92.4 & 92.1 & 92.0 & 92.6 & 92.2 \\
        &     w/o Fuzzy & 92.1 & 91.8 & 92.9 & 91.9 & 91.7 \\
        &     w/o C & 93.8 & 92.7 & 93.4 & 92.5 & 93.4 \\
        &     w/o M & 93.7 & 92.5 & 93.3 & 93.0 & 93.1 \\
        &     w/o F & 90.3 & 90.5 & 89.8 & 90.0 & 90.3 \\
        &     M-ADD & 90.3 & 89.8 & 88.4 & 90.1 & 90.0 \\
        &     M-CA & 95.9 & 95.8 & 95.7 & 95.5 & 95.5 \\
        &     \textbf{FMDNN} & \textbf{98.1} & \textbf{98.0} & \textbf{98.6} & \textbf{98.2} & \textbf{98.1} \\
        \bottomrule
    \end{tabular}
\end{table}



\begin{figure*}[!t]
\centering
\includegraphics[width=5.5in]{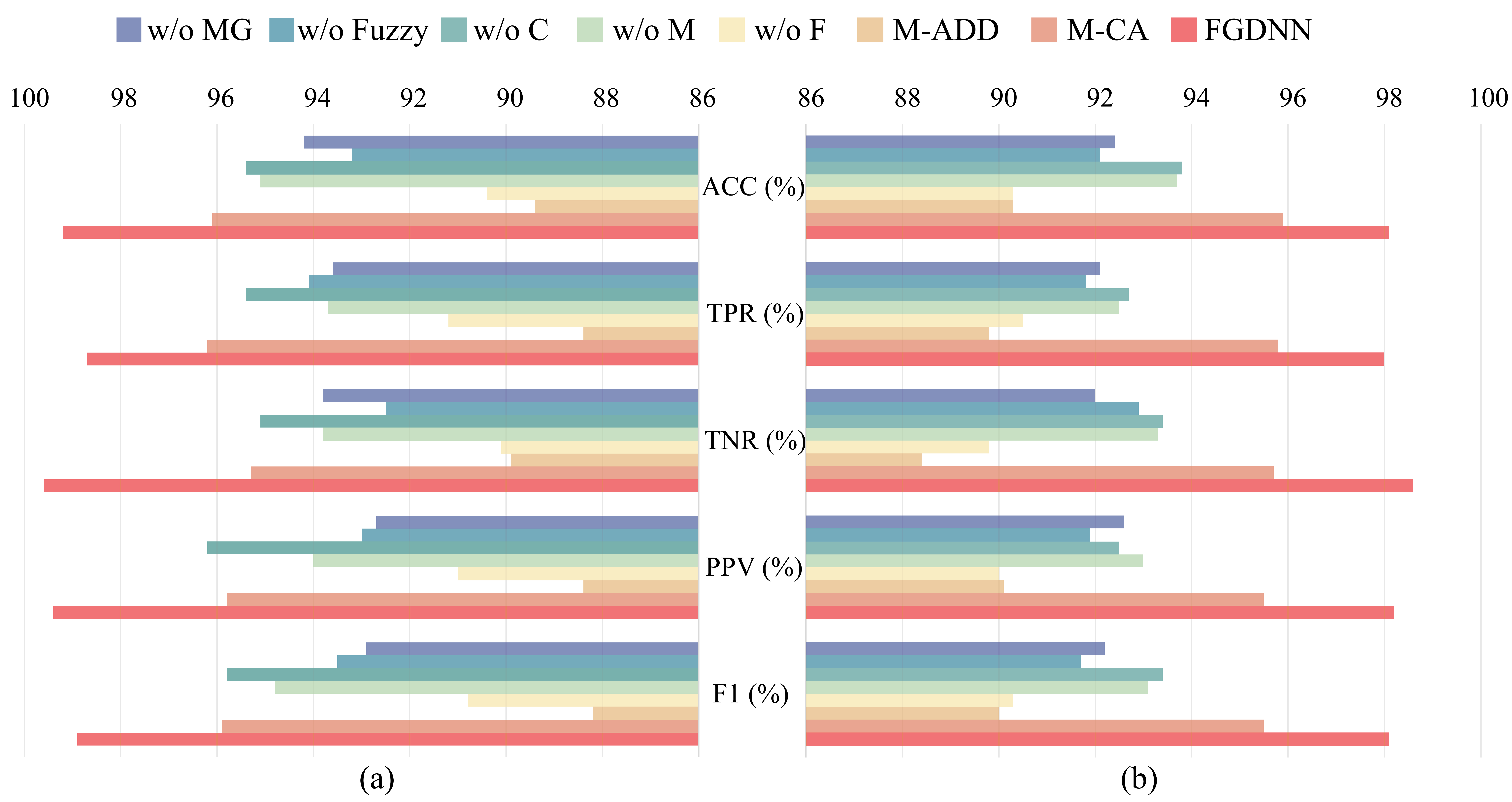}
\caption{Comparison of ablation experiments on different datasets: (a) LC; (b) NCT.}
\label{pic5}
\end{figure*}

\subsection{Ablation Experiment}
We conducted ablation experiments on FMDNN, using the LC and NCT datasets. We introduced specific modifications to the model, with experimental outcomes presented in Table IV and Fig. 5.

To verify the Multi-granularity Feature Extraction Module, we used a CNN without (w/o) multi-granularity, denoted as \textbf{w/o MG}, and eliminated coarse-, medium-, and fine-grained features, denoted as \textbf{w/o C}, \textbf{w/o M}, and \textbf{w/o F}, respectively.

To validate the Universal Fuzzy Feature Module, with all other conditions held constant, we replaced universal fuzzy features used as guiding information with features extracted by a standard CNN for model training, to compare the effectiveness of different guiding features. 
We call this condition \textbf{w/o Fuzzy}.

To verify the fuzzy-guided cross-attention module, while keeping all other conditions constant, we replaced fuzzy-guided cross-attention with regular additive fusion and standard cross-attention, respectively, denoted as \textbf{M-ADD} and \textbf{M-CA}. 

As shown in Table IV, the ACC of FMDNN shows that it significantly outperforms w/o MG by 5.0\% and 5.7\% on LC and NCT, respectively, and its F1-score is improved by 6.0\% and 5.9\%, respectively. These results demonstrate that multi-granular feature extraction enables the model to capture a broader range of feature information, thus enhancing its performance.

In w/o Fuzzy, there was a notable decrease in ACC by 6\% and 6\%, a reduction in TPR by 4.6\% and 6.2\%, and a decline in PPV by 6.4\% and 6.3\% on LC and NCT, respectively. Hence, we conclude that universal fuzzy features play a more pivotal role than multi-granular features. Histopathological images commonly suffer from interference from a substantial number of normal cells and influences from other tissues. Universal fuzzy features excel at filtering essential feature information from intricate histopathological images and demonstrate superior resilience to interference.

When training with features from two granularities, the evaluation metrics were consistently better than those with single-granularity features. On LC and NCT, fine- and medium-grained features w/o C exhibited F1-scores of 95.8\% and 93.4\%, respectively, and fine- and coarse-grained (w/o M) features showed F1-scores of 94.8\% and 93.1\%, respectively. Upon analysis, it is observed that fine-grained features capture more localized information, enabling the model to better capture subtle variations in the data, thereby improving model performance. However, when compared with F1-scores achieved by simultaneously extracting features from all three granularities, 98.9\% and 98.1\% respectively, a notable performance gap still exists.

We compared three fusion algorithms, as shown in Table IV. M-ADD, with ACC values of 89.4\% and 90.3\% and F1-scores of 88.2\% and 90.0\% on LC and NCT, respectively, did not exhibit significantly improved results after adding universal fuzzy features into multi-granular features. However, M-CA yielded slightly better results simply through the use of cross-attention. These findings highlight suboptimal performance due to the direct employment of additive fusion.


\begin{figure*}[!t]
\centering
\includegraphics{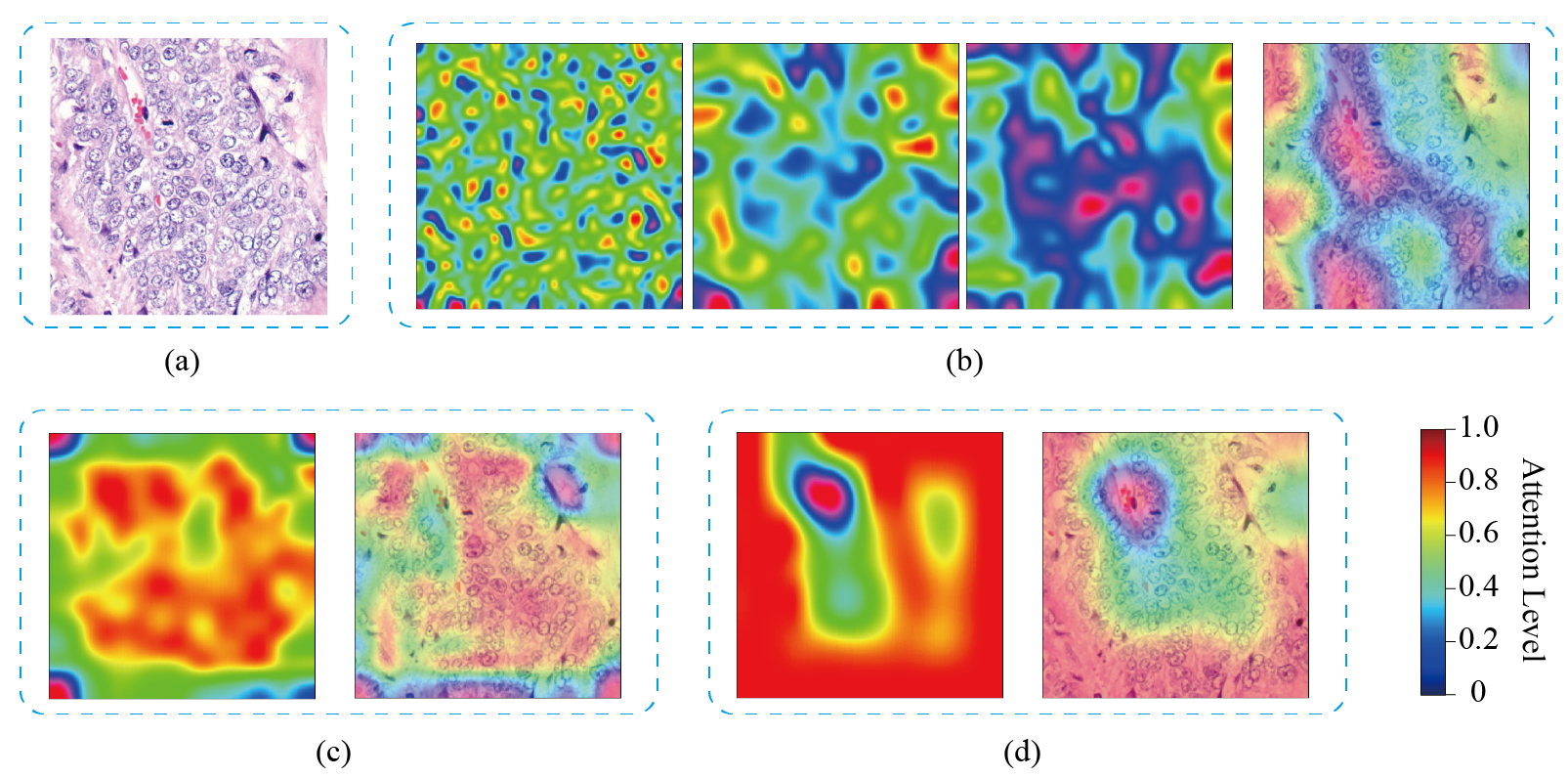}
\caption{Heatmap and CAM map of label "colonca:" (a) original image; (b) FMDNN fine-, medium-, and coarse-grained feature heatmap and CAM map in order; (c) EfficientNet feature heatmap and CAM map in order; (d) AlexNet feature heatmap and CAM map in order.}
\label{pic6}
\end{figure*}

\begin{figure*}[!t]
\centering
\includegraphics{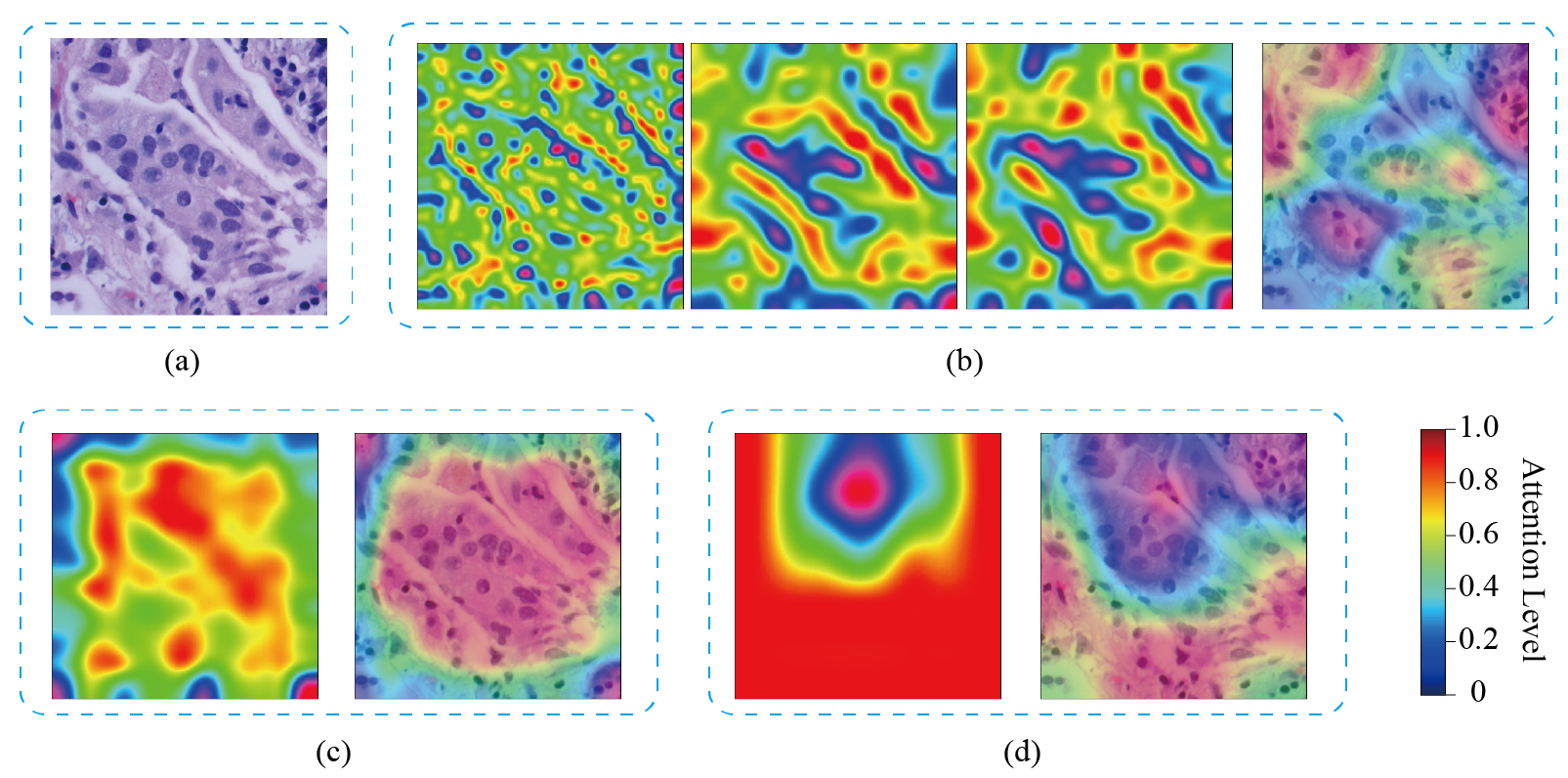}
\caption{Heatmap and CAM map of label "lungaca:" (a) original image; (b) FMDNN fine-, medium-, and coarse-grained feature heatmap and CAM map in order; (c) EfficientNet feature heatmap and CAM map in order; (d) AlexNet feature heatmap and CAM map in order.}
\label{pic7}
\end{figure*}


\subsection{Model Interpretability}
Feature visualization methods are employed to enhance the interpretability of multi-granular feature extraction. Heatmaps and Class Activation Maps (CAMs) are utilized to visualize the critical feature information learned by the model.

Heatmaps and CAM allow for a direct visual assessment of a model's attention across different spatial locations. We generated heatmaps for the three multi-granular features and compared them with other feature extraction methods.

  Figs. 6 and 7 show pathological images with respective "colonca" and "lungaca" classification labels as the subjects of analysis. Fig. 6(a) and 7(a) show the original image. In Fig. 6(b) and 7(b), from left to right, the sequence includes heatmaps and CAM maps for fine-, medium-, and coarse-grained features. From left to right, Fig. 6(c) and 7(c) show heatmaps and CAM maps derived from feature extraction by EfficientNet. In Fig. 6(d) and 7(d), from left to right, the sequence comprises heatmaps and CAM maps derived from feature extraction by AlexNet.

Observing the heatmaps in Figs. 6(c) and 7(c), the extracted features exhibit an indistinct separation between cells. This can be attributed to the complex network design, which limits interpretability in tasks involving fine granularity, similar categories, or subtle differences. The CAM maps indicate that the method focuses on patchy regions, with less emphasis on features in edge areas.

Examining the heatmaps in Figs. 6(d) and 7(d), the boundaries between cells are indistinct, and the separability is weak. This is attributed to the use of larger convolutional kernels in AlexNet, rendering it less sensitive in detecting smaller-granularity targets. In addition, the multifaceted nature of cells limits feature extraction to a single granularity, resulting in less accuracy and interpretability. The CAM maps further reveal that the model's focus extends over larger areas of the feature maps, making it susceptible to interference from non-critical features.

In Figs. 6(b) and 7(b), multi-granular feature heatmaps extracted by our model can better explain the multi-granularity attributes of cells. The proposed approach directs its attention at different granularities. At a coarse-grained level, it focuses on the overall cell layout, including spatial arrangement and aggregation; at a medium-grained level, it pays attention to the size and arrangement of cell populations; and at a fine-grained level, it hones in on details such as the size, shape, and chromatin distribution within cell nuclei. In addition, the CAM maps provide a visual representation of the network's activation regions in the images. FMDNN effectively and accurately captures features in pathological images, significantly enhancing classification accuracy.

\subsection{Limitations and Future Work}

Fig. 8 shows the F1-scores of FMDNN on datasets with different categories and sample sizes. The model achieves the expected results for datasets with a large number of samples, such as LC and NCT. The model performed commendably on datasets with less prominent multi-granular attributes, such as Bl, Kv, and especially HAM. However, there are still some limitations.

In reality, FMDNN is limited in the following ways: 1) The multi-granular feature extraction module exhibits suboptimal performance on datasets lacking discernible multi-granularity attributes as revealed by heat map analysis. 2) Our model demonstrates limited generalization in images with indistinct multi-granularity attributes. Employing NCT's weights on BL, HAM, and Kv yields ACCs of 71.2\%, 50.6\%, and 68.4\%, respectively. The model has not yet acquired sufficient semantic information to optimize its performance effectively for these specific images.

We consider the following potential solutions to these limitations: 1) A dynamic gate-controlled multi-granular FMDNN could be designed that optimizes various granular features through a gate mechanism. 2) In the future, CLIP\cite{ref68} could be integrated into the algorithm to enhance its semantic learning capabilities by maximizing the similarity of relevant sample pairs and learning model parameters.

\begin{figure}[!t]
\centering
\includegraphics[width=3.4in]{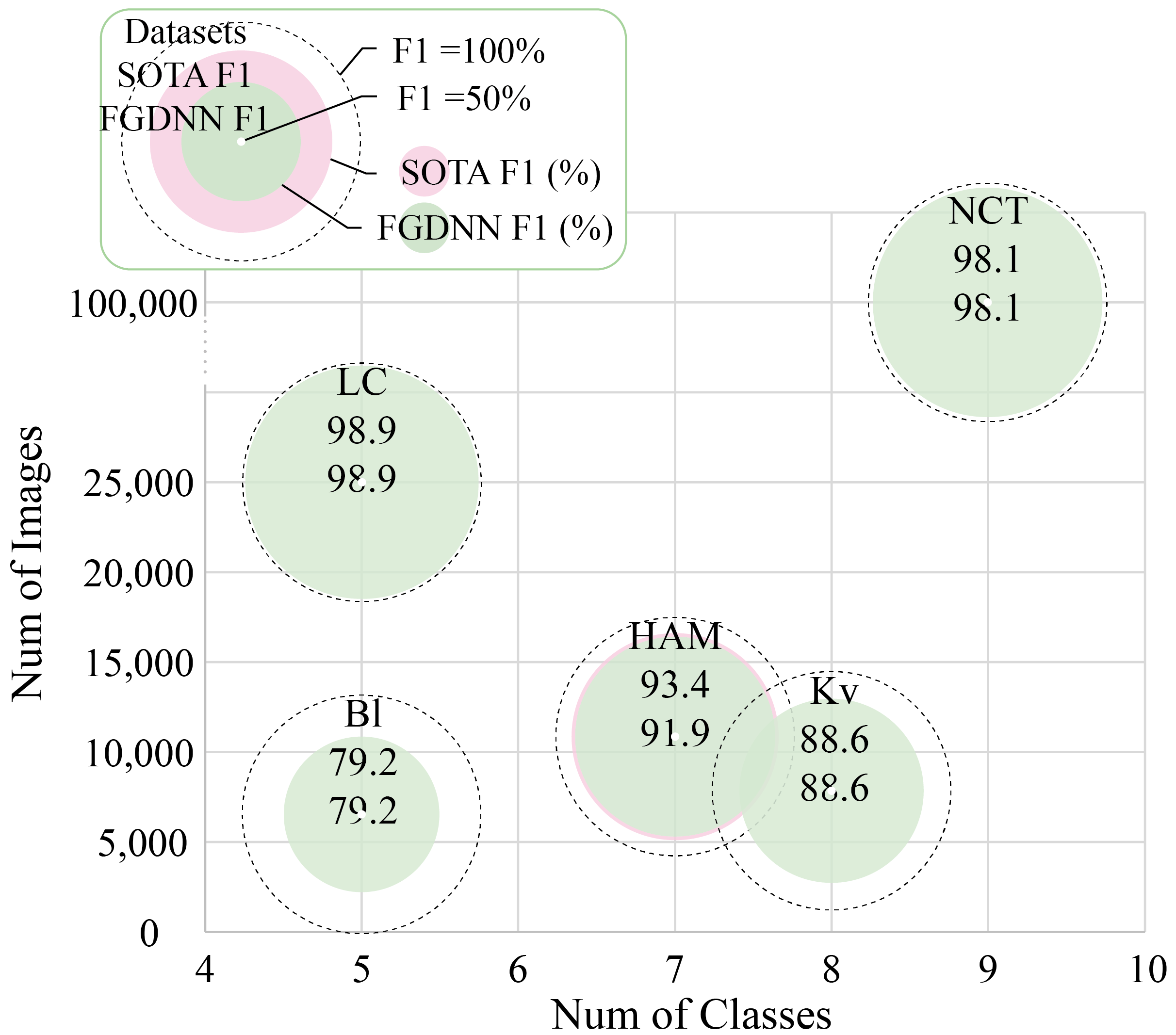}
\caption{FMDNN performance.}
\label{pic8}
\end{figure}

\section{Conclusion}
We introduced a Fuzzy-guided Multi-granularity Deep Neural Network (FMDNN), leveraging the unique multi-granularity attributes of cells to extract multi-granular features from histopathological images for improved information representation. Classical fuzzy set theory was introduced to address the challenge of information redundancy in multi-input scenarios. Fuzzy membership functions were used to extract general features from pathological tissue images. Fuzzy-guided cross-attention was employed to fuse critical feature information. Guided by the universal fuzzy features, the model showed enhanced generalization and robustness on datasets where multi-granularity attributes are obvious. Experimental evaluations on five publicly available datasets and comparative analyses with other models demonstrated the superior classification accuracy of FMDNN, particularly in tasks involving cellular interference and subtle features.

In our future work, we plan to adjust the resolution of feature extraction to enable better adaption to different types of datasets, especially to those with less pronounced multi-granularity attributes, to enhance the algorithm's generalizability and adaptability to a broad range of applications.
Given the relatively small size of histopathological image datasets, we will explore adversarial learning and transfer learning, improve the application of large models in medical datasets, and potentially include multimodal data to construct models with stronger generalization capabilities.

\bibliographystyle{IEEEtran}

\bibliography{reference.bib}

\newpage

\vfill

\end{document}